\documentclass[sigconf]{acmart}
\AtBeginDocument{%
  }

\copyrightyear{2026}
\acmYear{2026}
\setcopyright{cc}
\setcctype{by}
\acmConference[ICMR '26]{International Conference on Multimedia Retrieval}{June 16--19, 2026}{Amsterdam, Netherlands}
\acmBooktitle{International Conference on Multimedia Retrieval (ICMR '26), June 16--19, 2026, Amsterdam, Netherlands}
\acmDOI{10.1145/3805622.3810813}
\acmISBN{979-8-4007-2617-0/2026/06}

\usepackage{hyperref}
\begin{document}

\title{Revealing the Impact of Visual Text Style on Attribute-based Descriptions Produced by Large Visual Language Models}


\author{Xiaomeng Wang}
\affiliation{%
  \institution{Radboud University}
  \city{Nijmegen}
  \country{Netherlands}}
\email{xiaomeng.wang@ru.nl}

\author{Martha Larson}
\affiliation{%
 \institution{Radboud University}
 \city{Nijmegen}
 \country{Netherlands}
}
\email{m.larson@cs.ru.nl}

\author{Zhengyu Zhao}
\affiliation{%
  \institution{Xi'an Jiaotong University}
  \city{Xi'an}
  \country{China}
}
\email{zhengyu.zhao@xjtu.edu.cn}

\renewcommand{\shortauthors}{Xiaomeng Wang et al.}

\begin{abstract}
When the visual style of text is considered, a wide variety can be observed in font, color, and size.
However, when a word is read, its meaning is independent of the style in which it has been written or rendered.
In this paper, we investigate whether, and how, the style in which a word is visualized in an image impacts the description that a Large Visual Language Model (LVLM) provides for the concept to which that word refers.
Specifically, we investigate how functional text styles (readability-oriented, e.g., black sans-serif) versus decorative styles (display-oriented, e.g., colored cursive/script) affect LVLMs' descriptions of a concept in terms of the attributes of that concept.
Our experiments study the situation in which the LVLM is able to correctly identify the concept referred to by a visual text, i.e., by a word or words rendered as an image, and in which the visual text style should not influence the attribute-based description that the LVLM produces.
Our experimental results reveal that even when the concept is correctly identified, text style influences the model's attribute-based descriptions of the concept.
Our findings demonstrate non-trivial style leakage from text style into semantic inference and motivate style-aware evaluation and mitigation for LVLM-based multimedia systems.
\end{abstract}




\maketitle

\section{Introduction}
Large Visual Language Models (LVLMs) have recently exhibited competitive Optical Character Recognition (OCR) ability on visual text, i.e., words rendered in or as images.
\begin{figure}[t]
    \centering
    \includegraphics[width=\linewidth]{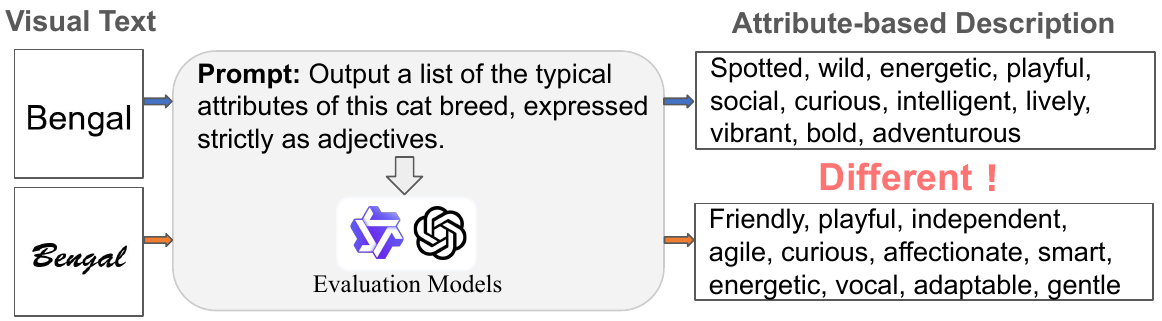}
    \caption{
    When a concept (left) is incorporated into a prompt (middle) in the form of visual text, the style of the concept impacts the attribute-description generated by the LVLM (right). 
    Top is \emph{functional text style} (here, Arial font) and bottom is \emph{decorative text style} (here, Brush font). }
    \label{fig:illustration}
\end{figure}
This capability is important for a range of applications of LVLMs to multimedia content in which the visual component includes text (e.g., user interfaces, memes, posters, and advertisements).
In this paper, we study whether, and how, the model is impacted by the style in which a word is written in search of insights on LVLMs' interpretation of visual text semantics, going beyond OCR performance.
Source code of our experiments is available online.\footnote{\url{https://github.com/XiaomengWang-AI/The-Impact-of-Visual-Text-style-on-Attribute-based-Descriptions-Produced-by-LVLMs}}

Our study considers the two contrasting text styles, which are illustrated in Figure~\ref{fig:illustration} for the example concept `Bengal', which is a spotted domestic cat breed, related to an Asian wild cat.
Functional text style (cf. top row) is a style used in cases in which the text must be readable.
In our experiments, functional text styles are represented by black, sans-serif visual texts.
Common sans-serif fonts are Arial, Calibri, and Consolas.
Decorative text styles (cf. bottom row) are used in cases where the text is displayed and needs to be readable, but is also intended to add something more. 
Decorative styles are primarily designed for display and aesthetic expression rather than continuous reading.
They often feature stylized strokes, script-like structures, and richer color usage (e.g., red, purple, and black) and emphasize visual flourish over readability.
Decorative styles are represented in our work by colored text and cursive or script fonts (e.g., Brush Script, Edwardian Script, Freestyle Script). 
Although, it is possible to imagine cases in which the style of the text is relevant for applications of LVLMs in multimedia analysis or retrieval, here, we study the case in which the style is not relevant and a correct response on the part of the LVLM would only use the meaning of the word and disregard the text style.

The focus of our work is attribute-based descriptions, namely, descriptions that an LVLM produces of a concept when it is prompted to generate a list of attributes describing that concept. 
We study attribute-based descriptions because they reflect the way in which the model captures the semantics of concepts.
As such, they are a good place to start in order to understand how LVLMs process text that is presented to them in images, i.e., in the visual modality, going beyond the question of OCR ability.
Our point of departure is that a robust model must ignore visual text style whenever it is not relevant.
Instability in the association of a visual text with the attributes of the concept that the text refers to could propagate to downstream tasks such as reasoning and retrieval, amplifying their impact on overall reliability. 

Our experiments study the patterns in attribute descriptions generated by LVLMs.
For each of our studied concepts, we collect two sets of attribute descriptions, one produced in response to visual text in functional style and one in response to visual text in decorative style.
Comparing the distributions of the attributes in these sets, we find a gap between identification and attribute inference: even when models correctly identify the concept written as visual text, their attribute-based descriptions are not style-invariant. 

\section{Related Work}
Recent work on LVLMs has demonstrated that models may respond differently when the same fact or concept is expressed as text versus depicted visually.
To quantify such failure in cross-model consistency, several benchmarks have been proposed.
PRISM~\cite{prism} constructs aligned question–answer instances spanning commonsense, factual, and mathematical queries in both text and image, and suggests that LVLMs often perform substantially better when the query is delivered in the text modality than when the same content is rendered visually.
REST/REST+~\cite{rest} further tests modality invariance by creating equivalent pairs of text and image or text and visual text, demonstrating that closing the OCR gap is insufficient to solve cross-modal inconsistency.
POPVQA~\cite{acl_entity_knowledge} finds substantial factual performance drops under visual reference conditioned on correct recognition, suggesting a late-stage visual-to-text information-flow bottleneck. 
In parallel,~\cite{emnlp_factual} reports degradation when recalling factual associations from visual and textual references, and shows that such failures correlate with distinctive internal states that can be detected by lightweight probes.
In this paper, we study inconsistencies when text is presented in the visual modality, thus exposing an overlooked axis of intra-modal inconsistency in LVLMs.

\section{Evaluating the Impact of Visual Text Style}
\label{sec:impact}
In this section, we describe the method by which we evaluate the impact of visual text style on LVLMs.
For each concept in a set of concepts, we create images in which that concept has been visualized as visual text, both in functional style and in decorative style.
Then, we apply visual text identification in order to create sets of visual texts that have been filtered such that they only contain visual texts for which the LVLM can successfully perform OCR. 
Using prompts that include the visual texts (as in Figure~\ref{fig:illustration}), we collect sets of attribute-based descriptions from the LVLMs.
Finally, we carry out a comparison of the distribution of attributes in the descriptions between functional and decorative styles.
In this section, we provide more details on how the attribute-based descriptions are collected and how the attribute distributions are analyzed.

\subsection{Collecting Attribute-based Descriptions}
We collect attribute-based descriptions by passing the model a visual text and text prompt that requests the model to produce a list of typical attributes (e.g., \emph{``Output a list of the typical attributes of this cat breed, expressed strictly as adjectives.''}).
The prompt is designed so that only the identity of the concept (rather than the style) written in the visual text is relevant to the answer.

For our investigation, we require a set of attribute-based descriptions that is representative of the general response of the LVLM in the case of visual text written in a specific style (functional or decorative).
Our exploratory experiments identified three major sources of influence, which we control for by including variation in our attribute-based description collection process, as follows:
(i) Prompt variation reduces prompt-induced bias~\cite{prompt_bias}: we use five semantically equivalent prompts with different phrasings.
(ii) Run repetition mitigates incidental variability: we prompt the model multiple times for each prompt (visual image + text prompt pair) with fresh, context-free runs.
(iii) Rendering variation addresses variability due to font size and spatial position: we define a set of configurations that covers the most influential differences.

\subsection{Comparing Attribute-based Descriptions}
Our procedure for collecting attribute-based descriptions for LVLMs yields, for each concept, a set of attribute lists for the functional case and a set of attribute-lists for the decorative case.
For each set, we merge the attribute lists contained in that set, resulting in a text that consists of terms (adjectives and phrasal descriptors) that are representative of what the LVLM produces for a given concept and a given style.
Over the terms in these texts, we can then calculate an empirical distribution.
For each concept, we first create a shared vocabulary $\mathcal{V}$ of terms by calculating the union of all terms occurring in either the functional or decorative texts.
Let $P$ and $Q$ denote the two distributions for the two styles, defined on the shared vocabulary of terms, $\mathcal{V}$.
We quantify the overall distribution difference using \emph{Total Variation} (TV) distance: $\mathrm{TV}(P,Q) \;=\; \frac{1}{2}\sum_{w \in \mathcal{V}} \left| P(w) - Q(w) \right|$, where $w$ is term in the vocabulary.
The range is from 0 to 1; larger values indicate larger distribution differences.

We conduct a Pearson's chi-squared test for homogeneity on the term counts to assess whether the observed distributional differences between visual text styles are statistically significant.
Specifically, for each concept, we construct a ($2\times K$) contingency table whose two rows correspond to the two styles and whose $K$ columns correspond to the terms in the $\mathcal{V}$ of that concept. 
To mitigate sparsity, we merge low-frequency adjective words (defined by a total count ($<\tau$) across both styles) into a bin. 
We evaluate the null hypothesis that the two styles induce the same attribute distribution for a given concept, and report the resulting $p$-values. 

We also examine the nature of the difference between the functional and the decorative distributions with an analysis of the top-most frequent attributes for each concept in the functional and in the decorative texts.
For every concept, we construct two lists of attributes ranked by frequency, one for the functional case and one for the decorative case.
We examine which attributes appear in top-$N$ attributes in one style list, but not the other.

\section{Experiments}
In this section, we specify the details of how we implement our evaluation of the impact of visual text styles, which was presented in Section~\ref{sec:impact}, and the results of our study.
\subsection{Experimental Setup}
\textbf{Evaluation models and dataset.}
We evaluate one open-source LVLM, Qwen2.5-VL-3B-Instruct\footnote{https://huggingface.co/Qwen/Qwen2.5-VL-3B-Instruct} and one closed-source LVLM, GPT-4o-mini\footnote{https://developers.openai.com/api/docs/models/gpt-4o-mini}.
For all queries, we set the temperature to 0. 
For Qwen, we run inference with $torch\_dtype=\texttt{torch.bfloat16}$.
We use the Oxford-IIIT Pet dataset~\cite{oxfordpets}, which contains labeled images of cats and dogs of different breeds.
Each class label is treated as a distinct concept.
We create visual text images by rendering each class label on an otherwise blank background. 

\noindent\textbf{Visual text generation.}
As summarized in Table~\ref{tab:style_settings}, for the functional style, we render the word in black using a sans-serif font category, and follow~\cite{font_recognition} to select {eight regular (non-italic, non-bold) fonts: Arial, Calibri, Consolas, Helvetica, Futura, Vera, Verdana, and Gill Sans.}
For the decorative style, we use eight script and cursive fonts: Brush, Edwardian, Freestyle, French, Magneto, Lucida
Handwriting, Mistral, and Segoe, and randomly choose one of five colors (black, red, blue, green, purple) for each rendering.
All fonts are downloaded as \texttt{.ttf} files.
We vary text size in {$\{30,35,40\}$} and consider three canonical placements: center, top-center, and bottom-center.
To cover common layouts, we use five size-position combinations: center at sizes {30/35/40}, and top-center/bottom-center at size {35}.
Overall, for each concept under each style condition, we generate 40 images (8 fonts $\times$ 5 size-position combinations).

\noindent\textbf{Filtering by identification performance.} 
We first test whether a {visual text} image can be correctly identified by the model in order to exclude cases in which OCR fails from our study of attribute-based descriptions.
For each image, we prompt the model to identify the concept:
\emph{``Identify the breed of the cat pictured in the image. Answer with the breed name directly.''}
(We replace \emph{cat} with \emph{dog} for dog breeds.)
The same prompt is used for both models and yields consistent responses upon re-prompting.
An image is counted as correctly identified only if the model output exactly matches the ground-truth class label (case-insensitive string match). 
This strict criterion provides a conservative estimate, as synonyms or partially correct names are counted as incorrect.

We filter out visual texts for which one of the two models makes an OCR mistake.
Then, we retain the breeds that have an adequate number of visual text images.
Choosing 32 breeds leaves us with at least 36 visual text images per concept and style.
(We eliminate `English Cocker Spaniel', `Newfoundland', `Birman', `Persian', and `Sphynx'.)
For the cases in which they are more than 36, we randomly sample to have exactly 36 visual text images.

\begin{table}
\caption{Settings of functional and decorative styles.}
\centering
\small
\setlength{\tabcolsep}{3pt}
\begin{tabular}{l p{2.55cm} p{2.55cm}}
\toprule
\textbf{Factor} & \textbf{Functional} & \textbf{Decorative} \\
\midrule
Font Category & Sans-serif & Cursive / script \\
Font &
{Arial, Calibri, Consolas, Helvetica, Futura, Vera, Verdana, Gill Sans} &
{Brush, Edwardian, Freestyle, French, Magneto, Lucida Handwriting, Mistral, Segoe} \\
Rendering color & Black & Uniform over \{black, red, blue, green, purple\} \\
Font size & \multicolumn{2}{c}{{\{30, 35, 40\}}} \\
Font position & \multicolumn{2}{c}{center / top-center / bottom-center} \\
Size-position combos & \multicolumn{2}{c}{{center@30/35/40}; {top/bottom-center@35}} \\
\bottomrule
\end{tabular}

\label{tab:style_settings}
\end{table}

\noindent\textbf{Collecting and processing attribute-based descriptions.} 
For the two sets of visual texts, functional and decorative, we collect the responses of the LVLMs.
We query each image with five prompts (cat version shown; \emph{cat} is replaced by \emph{dog} for dog breeds):
(1) \emph{``Output a list of the typical attributes of this cat breed, expressed strictly as adjectives.''}
(2) \emph{``Output a list of attributes that distinguish this cat breed from other cat breeds, expressed strictly as adjectives.''}
(3) \emph{``Output a list of adjectives that describe this cat breed.''}
(4) \emph{``Output a list of adjectives that capture how this cat breed is different from other cat breeds.''}
(5) \emph{``Produce a list of the typical characteristics of this cat breed, expressed strictly as adjectives.''}
The prompts were designed to vary with respect to these aspects: description vs. distinction, asking for adjectives/list, and substituting synonyms of keywords.
We confirmed the prompts produced actual lists of attributes (adjectives or descriptive phrases), and also tested to ensure they worked for both models.
We submit each prompt to each model five times and collect the five responses.
Each repetition is executed as an independent run with no context carried over between runs (i.e., the model is re-invoked from a fresh context each time).
The final set of collected LVLM responses consisted of, for each concept and each style, 900 lists of attribute outputs per model (36 visual texts $\times$ 5 prompts $\times$ 5 re-prompts). 

Note that the raw outputs of Qwen2.5-VL-3B-Instruct are not always format-consistent (e.g., mixed punctuation or additional explanations or sentence structure). 
To standardize evaluation, we use an auxiliary LLM, Llama-3.1-8B\footnote{https://huggingface.co/meta-llama/Llama-3.1-8B} to extract adjectives (or phrasal descriptors) and return a single comma-separated list.
We use the following system prompt:
\emph{``Act as a text processing system. Extract adjectives (or adjective phrases) from the input and output only a single-line list separated by comma. No other text.''}

\subsection{Experimental Results}
\textbf{Identification performance.}
Recall that our study uses only visual texts for which the LVLM is able to correctly identify the word, i.e., OCR is successful.
We calculate the visual identification performance on the basis of the 1480 original visual texts.
Qwen2.5-VL-3B-Instruct achieves accuracies of 0.9851 and 0.9601 under the functional and decorative settings, respectively, whereas GPT-4o-mini achieves 0.9885 and 0.9905.
These results confirm that overall the OCR ability of the LVLM is high and that by studying only correctly identified words we are actually studying a majority of the cases that can be expected to occur.

\noindent\textbf{Comparing attribute distributions.}
Here, we present our analysis of the impact of visual text style on attribute-based definitions.
We start by noting that the attributes produced by the LVLMs are reasonable, they often describe the temperament of dog and cat breeds, but also include their physical characteristics.  
For example, when the prompt includes the visual text `Chihuahua', the attribute words commonly are small-sized, short-haired, and active. 
In this work, we do not examine the correctness of the attribute-based descriptions of the breeds, but rather look at the impact of the visual text style on these descriptions.

\begin{figure}[!t]
    \centering
    \includegraphics[width=\linewidth]{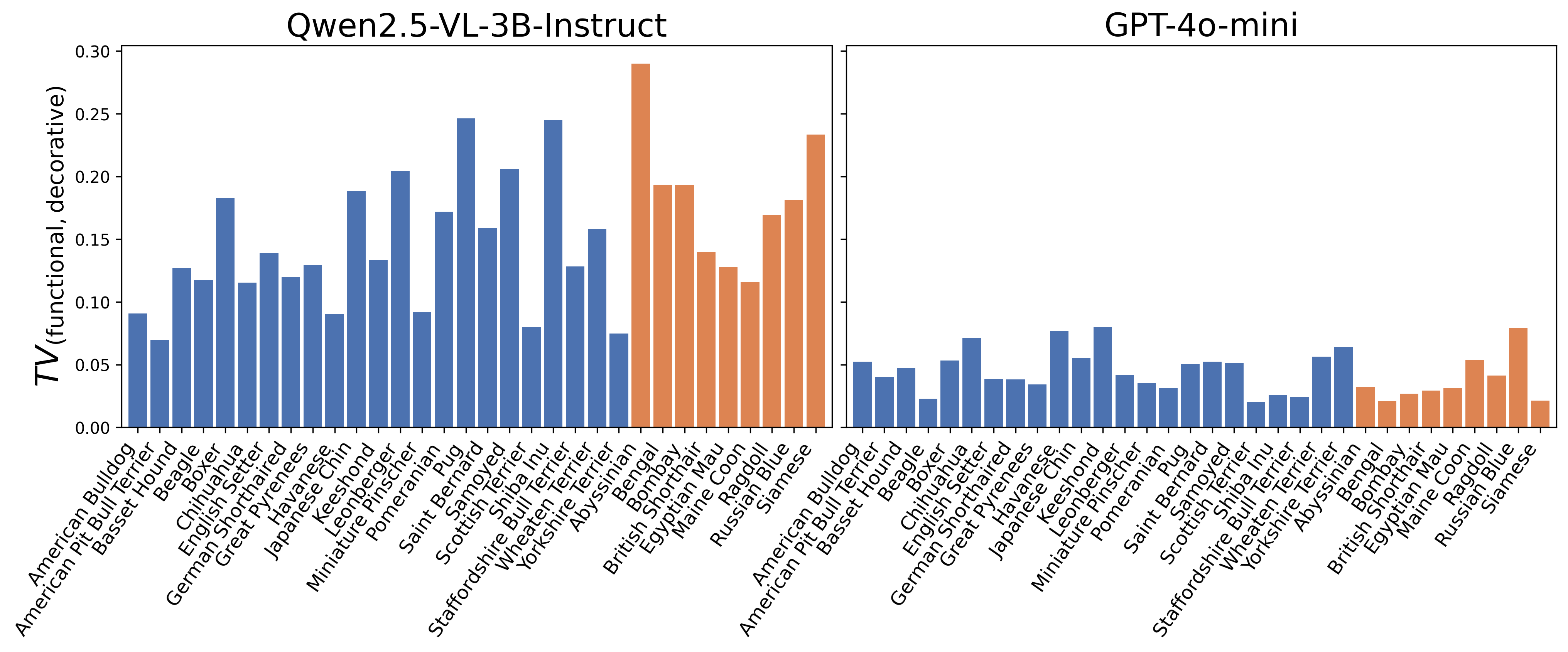}
    \caption{TV values between functional and decorative style distributions across breeds for Qwen2.5-VL-3B-Instruct and GPT-4o-mini. Blue and orange denote dog and cat breeds, respectively; breeds are alphabetically ordered.}
    \label{fig: tv}
\end{figure}

Results of our analysis are presented in Figure~\ref{fig: tv}.
We see that the TV values between functional and decorative styles are consistently quite different from zero, reflecting a shift in the distribution between visual text styles. 
The corresponding $p$-values indicate significant differences ($p<0.001$ for all cases).
For the $p$-value calculation, the threshold $\tau$, which controls merging of low-frequency terms, is set to 5.
Note that we have calculated the $V$ for each concept separately, which means across concepts the sample space on which $TV$ distance is calculated differs. 
For this reason, it is important not to over-interpret the exact differences in $TV$ distance between concepts.
However, assuming that the sample spaces are relatively comparable, then we can conclude that Qwen-2.5-VL-3B-Instruct exhibits greater style sensitivity, than GPT-4o-mini.

\begin{figure}[!t]
    \centering
    \includegraphics[width=\linewidth]{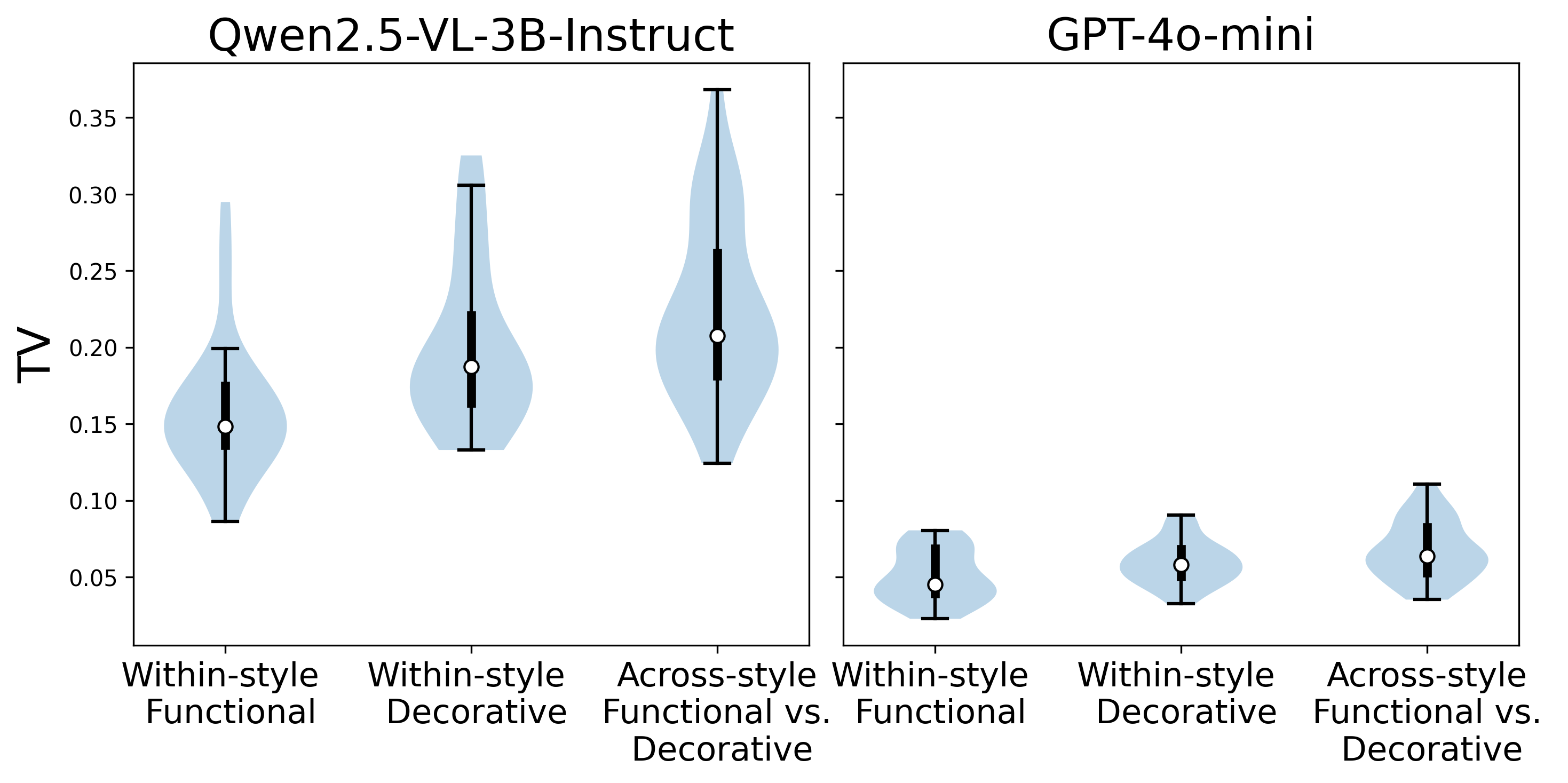}
    \caption{Average pairwise TV values show that across-style variation exceeds within-style variation for both models. 
    }
    \label{fig: tv_within_font}
\end{figure}

Next, we compare the within-style distance to the across-style distance calculated by comparing the attribute-based descriptions associated with the individual fonts.
For each font, a distribution is estimated over the set of attribute-based lists produced in response to visual texts written in that font.
The within-style distance is the average pairwise $TV$ distance of fonts associated with the same style and the across-style distance is the average pairwise $TV$ distance of fonts not associated with the same style.
As shown in Figure~\ref{fig: tv_within_font}, across-style distance consistently exceeds within-style distance for both models. 
This comparison confirms that the $TV$ distances in Figure~\ref{fig: tv} can be attributed to the difference between styles, and not to other factors, such as our specific choice of fonts or general variation in the vocabularies of the model.

Finally, we look at the nature of the difference between functional and decorative style by examining the differences in the top-$N$ most-frequent attributes produced in the two cases.
For our analysis, we take $N$ to be three,  
which for both models corresponds approximately to adjectives that occur in 75\% of the attribute output lists.
The top-3 attributes refer either to the temperament or the physical form of the breed. 
We confirm which attributes we should consider temperamental by consulting the Breed Temperament Guide issued by the American Kennel Club (AKC)\footnote{https://www.akc.org/akctemptest/breed-temperament-guide/}.

For both Qwen-2.5-VL-3B-Instruct and GPT-4o-mini, the top attributes are predominantly temperamental.
However, for breeds where a physical attribute appears among the top three, it is more likely to occur in the functional style case than in the decorative style case. 
Specifically, for Qwen-2.5-VL-3B-Instruct, there is a difference between the top-3 attributes for functional and for decorative in 13 of the 32 breeds.
Of these 13 breeds, there are five breeds for which this discrepancy involves a physical attribute that occurs in one style but not in the other, replacing a temperamental attribute:
\emph{compact} in place of \emph{versatile} (`Miniature Pinscher'), 
\emph{strong} in place of \emph{gentle} (`Maine Coon'), \emph{small} in place of \emph{loyal} (for `Pomeranian'), \emph{small} in place of \emph{playful} (for `Pug'), and \emph{fluffy} in place of \emph{affectionate} (for `Ragdoll').
In four out of these five cases, the physical attribute occurs in the functional top-3 attributes.
Only in the case of \emph{compact} (`Miniature Pinscher') does the physical attribute that is substituted for a temperamental attribute occur in the decorative top-3.
Note that in order to influence the top-3 attributes, which are calculated on sets of 900 LVLM, the impact of style on attribute-based descriptions output by the LVLM must be non-trivial, i.e., large enough to potentially be important in applications. 

\section{Conclusion and Outlook}
In this work, we have investigated the impact of the visual style in which a concept is written on the attribute descriptions that an LVLM produces for that concept.
Our analysis compared the distribution of attributes produced by LVLMs in response to functional and decorative visual text styles.
Our results reveal a substantial gap: models can often correctly identify concepts from rendered {visual text}s, yet their attribute predictions lack style invariance. 
Our work suggests that {visual text style} has an influence on how LVLMs capture the semantics of concepts.
The finding that the style can leak from visual text into semantic inference motivates, in future work, the incorporation of style variation factors into evaluation protocols for multimedia systems deployed in text-rich visual environments.

We conclude by noting our results suggest that, compared to functional text, decorative text may be more strongly connected to temperamental attributes.
Both share a connection to emotion: Breed temperament is important for the bond that people form with their pets, and decorative style is important for the emotional reaction to visual text. 
More work is needed to confirm whether visual text style indeed introduces an emotion-related bias.

\bibliographystyle{ACM-Reference-Format}
\bibliography{reference}










\end{document}